\documentclass[twoside,11pt]{article}

\usepackage{jmlr2e}
\usepackage{makecell}

\newcounter{daggerfootnote}
\newcommand*{\daggerfootnote}[1]{%
    \setcounter{daggerfootnote}{\value{footnote}}%
    \renewcommand*{\thefootnote}{\fnsymbol{footnote}}%
    \footnote[2]{#1}%
    \setcounter{footnote}{\value{daggerfootnote}}%
    \renewcommand*{\thefootnote}{\arabic{footnote}}%
    }

\ShortHeadings{Automated CXR Reading using Dual Convolutional Neural Networks}{Rubin, Sanghavi, Zhao, Lee, Qadir \& Xu-Wilson}
\firstpageno{1}

\begin{document}

\title{Large Scale Automated Reading of Frontal and Lateral Chest X-Rays using Dual Convolutional Neural Networks}

\author{\name Jonathan Rubin \email Jonathan.Rubin@philips.com \\
       \name Deepan Sanghavi \email Deepan.Sanghavi@philips.com \\
       \name Claire Zhao \email Claire.Zhao@philips.com \\
       \name Kathy Lee \email Kathy.Lee\_1@philips.com \\
       \name Ashequl Qadir \email Ashequl.Qadir@philips.com \\
       \name Minnan Xu-Wilson \email Minnan.Xu@philips.com \\
       \addr   Acute Care Solutions\\
       Philips Research North America\\
       Cambridge, MA, United States} 

\maketitle

\begin{abstract}
The MIMIC-CXR dataset is (to date) the largest released\daggerfootnote{At the time of writing, MIMIC-CXR has been made available as a limited release and is intended for dissemination in the near future.} chest x-ray dataset consisting of 473,064 chest x-rays and 206,574 radiology reports collected from 63,478 patients. We present the results of training and evaluating a collection of deep convolutional neural networks on this dataset to recognize multiple common thorax diseases. To the best of our knowledge, this is the first work that trains CNNs for this task on such a large collection of chest x-ray images, which is over four times the size of the largest previously released chest x-ray corpus (ChestX-Ray14). We describe and evaluate individual CNN models trained on frontal and lateral CXR view types. In addition, we present a novel DualNet architecture that emulates routine clinical practice by simultaneously processing both frontal and lateral CXR images obtained from a radiological exam. Our DualNet architecture shows improved performance in recognizing findings in CXR images when compared to applying separate baseline frontal and lateral classifiers.
\end{abstract}

\section{Introduction}

Automatically classifying findings of interest within chest radiographs remains a challenging task. Systems that can perform this task accurately have several use-cases. In particular, chest x-rays are the most commonly ordered imaging study for pulmonary disorders \citep{raoof2012interpretation} and given the sheer volume of images being produced, an automated system that provides secondary reads for radiologists could allow important findings not to be missed. Moreover, in regions where access to trained radiologists is limited, an automated system that can accurately detect thoracic diseases from chest x-rays would be greatly valuable. Alternatively, in fast-paced care settings such as the emergency department and intensive care unit, clinicians may not have time to wait for the results of a radiology report to become available. A system that can automatically flag potentially lethal conditions (e.g. complications from mechanical ventilation leading to pneumothorax \citep{chen2002pneumothorax}) could allow care providers to respond to emergency situations sooner.

\begin{figure}[htbp]
  \centering 
  \includegraphics[width=0.65\columnwidth]{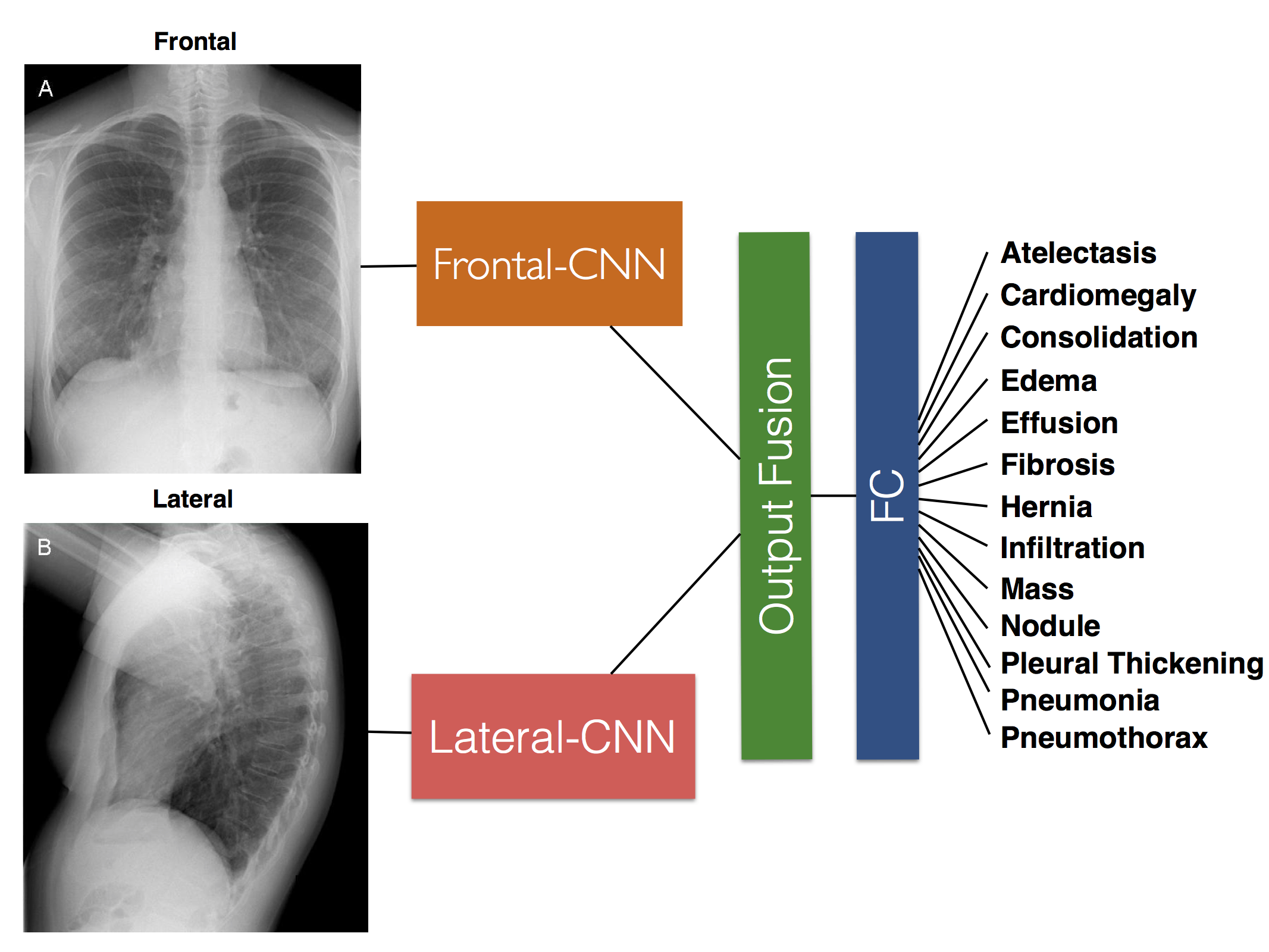} 
  \caption{The DualNet architecture accepts a pair of frontal (PA or AP) and lateral input images. Two convolutional neural networks are trained in parallel for each input and their outputs are concatenated together before a final fully connected layer makes a multi-label prediction.}
  \label{fig:dualnet} 
\end{figure} 

In this work, we train deep convolutional neural networks for the purpose of automatically classifying findings in chest x-ray images. We make use of the MIMIC-CXR dataset, which has been made available via a limited release. The MIMIC-CXR dataset is intended for future dissemination and consists of the largest available set of frontal and lateral chest radiographs, compared to previously released datasets. The dataset consists of 473,064 chest x-rays in DICOM format collected from 63,478 patients. In addition, 206,574 radiology reports that correspond to the CXR images are also available. Consistent with recent previous works on automated chest x-ray analysis \citep{yao2017learning,rajpurkar2017chexnet,guan2018diagnose,kumar2017boosted,baltruschat2018comparison}, we focus on recognizing 14 thoracic disease categories, including {\it atelectasis, cardiomegaly, consolidation, edema, effusion, emphysema, fibrosis, hernia, infiltration, mass, nodule, pleural thickening, pneumonia} and {\it pneumothorax}. Unique from previous works, we train separate models based on the view position of the radiograph. In particular, separate CNN models are trained for posteroanterior (PA), anteroposterior (AP) and lateral view position CXR images. Furthermore, consistent with how standard chest examinations take place \citep{virginiamedicine}, we present a novel DualNet architecture that accepts as input both a frontal and lateral chest x-ray taken from a patient during a radiographic study. An overview of the DualNet architecture is graphically depicted in Fig.~\ref{fig:dualnet}. Both the frontal and lateral CXR inputs are processed by separate convolutional neural networks and their outputs are combined into a fully connected layer to make a final classification. We compare the DualNet architecture to baseline CNN architectures that process PA, AP and lateral inputs separately and show that processing both frontal and lateral inputs simultaneously leads to improvement in classification performance.

\section{Related Work}

While computer-aided diagnosis in chest radiography has been studied for many years \citep{ginneken2001computer,van2001computer}, the public release of the ChestX-ray14 dataset \citep{wang2017chestx} has resulted in many recent works that attempt to classify thorax diseases from frontal chest x-rays \citep{yao2017learning,rajpurkar2017chexnet,guan2018diagnose,kumar2017boosted,baltruschat2018comparison}. Perhaps the most well-known of these works is that of \cite{rajpurkar2017chexnet}, in which the authors present the ChexNet system. In addition to evaluating ChexNet by determining how well the model classifies 14 thorax diseases labeled within the ChestX-ray14 dataset, \cite{rajpurkar2017chexnet} extracted a subset of 420 images and evaluated the performance of a binary ChexNet model trained to detect pneumonia. They compare the performance of ChexNet to that of four practicing academic radiologists and show that ChexNet was able to achieve a better $F_1$ score compared to the average score from the radiologists.

A recent work by \citep{guan2018diagnose} uses the ChestX-ray14 dataset to train attention-guided networks (AG-CNN) which consist of a standard global branch CNN that processes a full CXR image, as well as a local branch CNN that processes the result of applying a CAM-like saliency map \citep{zhou2016learning} for region localization. When both branches are fine-tuned together they show improved performance over utilizing each branch individually. The authors of \citep{guan2018diagnose} present improved results over those presented by \citep{rajpurkar2017chexnet}, however, they perform a random 70/10/20 train/validation/test set split that does not ensure the same subject is not present in different sets.

Apart from the ChestX-ray14 dataset, previously smaller publicly released datasets have also been used to construct CXR classifiers. The Japanese Society of Radiological Technology (JSRT) dataset \citep{shiraishi2000development} is a small frontal CXR dataset that contains normal images, as well as radiographs exhibiting malignant and benign lung nodules. In \citep{gordienko2017deep}, the authors train small convolutional neural networks both with and without lung segmentation for the task of classifying whether lung nodules are present within a CXR image or not. They apply their method on the JSRT dataset, as well as on a version of the JSRT dataset where bone shadows have been eliminated, BSE-JSRT \citep{van2006segmentation}. Due to the limited size of the JSRT dataset (247 images), the training/validation set results presented in \citep{gordienko2017deep} exhibit significant overfitting, however the authors do show improved performance when using the BSE-JSRT dataset with bone shadows removed.

Finally, the Indiana chest X-ray collection consists of 8,121 images and 3,996 corresponding radiology reports \citep{demner2015preparing}. In \citep{islam2017abnormality}, the authors combined the Indiana chest X-ray collection with the JSRT dataset, as well as the Shenzhen dataset \citep{jaeger2014two}. They compare the performance of several CNN architectures such as AlexNet \citep{krizhevsky2012imagenet}, VGG \citep{simonyan2014very} and ResNet \citep{he2015deep} and show that an ensemble of models leads to classification performance improvement.

While a wide range of works continue to become available that apply deep convolutional networks to chest x-rays images, a few common drawbacks exist that limit their applicability for constructing automated CXR classifiers. In particular, approaches trained on the ChestX-ray14 dataset accept 8-bit, grayscale PNG images as model inputs. This is a limited dynamic range compared to DICOM formats that typically encode medical image pixel and voxel data using 12-bit depth, or greater. In addition, the released PNG images in the ChestX-ray14 dataset were resized to 1024x1024 pixel width and height. This resizing was performed without maintaining the original aspect ratio and could introduce distortion into the images. Several of the works severely downsample input images to match dimensions required for pre-trained classifiers \citep{rajpurkar2017chexnet,guan2018diagnose}. Moreover, the mentioned recent works \citep{yao2017learning,rajpurkar2017chexnet,guan2018diagnose,kumar2017boosted,baltruschat2018comparison} make no distinction between PA and AP chest x-rays view positions. This can be a problem for some findings, such as cardiomegaly, which can only be accurately assessed in PA images, as the AP view will exaggerate the heart silhouette due to magnification. Finally, the above works focus solely on the evaluation of frontal chest x-rays, whereas the lateral view reveals lung areas that are hidden in the frontal view \citep{raoof2012interpretation}. The lateral view can be especially useful in detecting lower-lobe lung disease, pleural effusions, and anterior mediastinal masses \citep{auntminnie2001positioning} and is hence routinely taken into account by practicing radiologists.

\section{Contributions}

The contributions of this work are as follows:
\begin{enumerate}
\item We train deep convolutional neural networks to recognize multiple common thorax diseases on the as-yet largest collection of chest radiographs -- the MIMIC-CXR dataset.
\item We describe and evaluate CNN models for processing frontal, as well as lateral chest x-rays, which have received less attention from previous research efforts. Furthermore, we develop distinct models for anteroposterior and posteroanterior frontal view types.
\item We introduce a novel DualNet architecture, which simultaneously processes a patient's frontal and lateral chest x-rays and demonstrate its usefulness in improving performance against baseline classifiers.
\end{enumerate}

\section{Model architectures}

To classify thorax diseases in chest x-ray inputs, we train separate models based on the type of view position that was used to acquire the image. In particular, three networks are trained specifically for posteroanterior (PA), anteroposterior (AP) and lateral (Lateral) view types. Furthermore, we introduce a new network architecture, referred to as DualNet, that accepts paired frontal and lateral CXR inputs. We train two types of DualNet architectures, one for PA-Lateral pairs and one for AP-Lateral pairs. A schematic of the DualNet architecture is shown in Fig.~\ref{fig:dualnet}.

For each model type, the baseline CNN architecture used is a modified version of DenseNet-121 \citep{DBLP:conf/cvpr/HuangLMW17}. The original DenseNet-121 model is altered by replacing the typical 3-channel (RGB) input layer with 1-channel (grayscale) input. Four DenseNet blocks are applied using a growth rate of 32 channels per layer. 

To allow the handling of different image input sizes, we perform a global average pooling operation on the final convolutional layer before a fully connected layer maps 1x1 feature maps to 14 output classes. As each input image could potentially contain multiple findings the task of classifying thorax diseases in chest x-rays is treated as a multi-class, multi-label problem. As such, a sigmoid operation is applied to each of the 14 outputs and a binary cross-entropy loss function is used to train each network.

\section{Data processing}

\subsection{Data splitting}

The MIMIC-CXR dataset was split into separate training, validation and test sets. 20\% of the images were held out as test data. The remaining 80\% of the data was then split between the training (70\%) and validation (10\%) sets. The dataset was split by subject to ensure a distinct set of subjects existed in training, validation and test datasets, i.e. no subject was present in more than one dataset group. The data splits were then verified to ensure prevalence between view types and thoracic disease categories were consistent between groups. Table \ref{tab:split_totals} shows the overall number of training, validation and test set images used to train and evaluate separate PA, AP and lateral view models. In addition, Table \ref{tab:dual_totals} depicts the number of radiological studies where frontal and lateral pairs of images were available to train PA-Lateral and AP-Lateral DualNet models.

\begin{table}[htbp]
  \centering 
  \caption{Total images per split} 
  \begin{tabular}{|l|l|l|l|}
    \hline
    & Train & Validation & Test\\
    \hline
    PA & {\it 70,875} & {\it 10,078} & {\it 20,426}\\
    \hline 
    AP & {\it 168,525} & {\it 24,566} & {\it 47,689}\\
    \hline 
    Lateral & {\it 81,379} & {\it 11,519} & {\it 23,125}\\
    \hline 
    {\bf Total} & {\bf 320,779} & {\bf 46,163} & {\bf 91,240}\\
    \hline
  \end{tabular}
  \label{tab:split_totals}
\end{table}

\begin{table}[htbp]
  \centering 
  \caption{Number of radiological studies with frontal and lateral dual pairs} 
  \begin{tabular}{|l|l|l|l|}
    \hline
    & Train & Validation & Test\\
    \hline
    PA \& Lateral & {\it 59,500} & {\it 8,414} & {\it 17,146}\\
    \hline 
    AP \& Lateral & {\it 15,372} & {\it 2,186} & {\it 4,257}\\
    \hline
    {\bf Total} & {\bf 74,872} & {\bf 10,600} & {\bf 21,403}\\
    \hline
  \end{tabular}
  \label{tab:dual_totals} 
\end{table}

\subsection{Data transformations}

A series of transformation operations were applied to the data before model training. First, nearest-neighbor interpolation was used to scale images to a specified dimension, while ensuring original aspect ratio was maintained. Images were then cropped to enforce equivalent width and height. In our experiments, we used an image width and height of 512 x 512 pixels. Pixels were then normalized to between 0 and 1 by dividing by the maximum grayscale value $2^{12}-1$.

\subsection{Instance Labeling}

Chest x-ray images were labeled from their corresponding radiology reports using NegBio \citep{peng2017negbio,wang2017chestx}. NegBio maps each report into a collection of Unified Medical Language System (UMLS) concept ids. Images were first mapped to an initial set of 46 UMLS concept ids. A further mapping took place to subsequently assign these 46 concepts to the 14 common thoracic disease types commonly used in previous works \citep{wang2017chestx,rajpurkar2017chexnet}. As such, more specific concept ids were mapped to their more general class, e.g. C0546333 (right pneumothorax) and C0546334 (left pneumothorax), were both mapped to the general concept - pneumothorax. (Note: emphysema did not occur in the MIMIC-CXR dataset, so was excluded, giving a total of 13 CXR findings). 

To assign final labels, only annotations from the `Findings' and `Impression' sections of radiology reports were used. Furthermore, a positive label was only assigned if the concept was not identified as being negated or uncertain by NegBio. An individual image could be assigned multiple positive labels for each of the 13 common thoracic disease types. Images where none of the 13 disease types were identified were labeled as `No Finding'. Table \ref{tab:prevalence} presents the ordered per class prevalence for each of the final 14 class labels.

\begin{table}[htbb]
  \centering 
  \caption{Overall prevalence per class. Note: a single image may contain multiple conditions (percentages shown are out of total positive labels).}
  \begin{tabular}{|l|l|r|}
    \hline
     & Total & Percentage\\
    \hline
    No Finding			& 200,137		& 30.57\%\\
    Effusion			& 123,485		& 18.86\%\\
    Atelectasis			& 105,815		& 16.16\%\\
    Cardiomegaly		& 67,446		& 10.30\%\\
    Edema				& 52,597		& 8.03\%\\
    Consolidation		& 29,494		& 4.51\%\\
    Pneumonia			& 25,025		& 3.82\%\\
    Pneumothorax		& 23,551		& 3.60\%\\
    Nodule				& 6,972		& 1.07\%\\
    Pleural Thickening		& 5,696		& 0.87\%\\
    Hernia				& 4,930		& 0.75\%\\
    Infiltration			& 4,766		& 0.73\%\\
    Mass				& 3,381		& 0.52\%\\
    Fibrosis		 		& 1,365 		& 0.21\%\\
    \hline
  \end{tabular}
  \label{tab:prevalence} 
\end{table}

\section{Model training}

Baseline PA, AP and Lateral models were seeded using pre-trained ImageNet \citep{ILSVRC15} weights. No data augmentation took place and Adam optimization \citep{kingma2014adam} was used with a cyclic learning rate \citep{smith2017cyclical}. An initial learning rate range test established a learning rate boundary of between 0.001 -- 0.02 in which the learning rate then fluctuates between during training. The Triangular2 policy \citep{smith2017cyclical} was used to control how learning rate fluctuations are altered over time. Finally, stratified mini-batch sampling was employed to ensure each mini-batch maintained overall class prevalence during training. PyTorch was used for model development and models were trained using data parallelism over 8 Nvidia Titan Xp GPUs.

\section{Results}

\subsection{Frontal and Lateral Model Evaluation}

Table \ref{tab:results1} presents the per class AUC results calculated on the held-out test set for each of the PA, AP and Lateral models described. It can be seen that recognition performance for each of the findings varies by view type. 
Compared with AP and lateral views, PA models result in larger AUC values for {\it atelectasis}, {\it cardiomegaly}, {\it fibrosis}, {\it infiltrates} and {\it pleural thickening}. For frontal view types, the PA model achieves a larger average AUC ({\it 0.702}), compared to the AP model ({\it 0.655}). This difference is likely due to changes in the clinical setting where these images are acquired. In particular, AP images are typically obtained in the intensive care unit. Apart from heart and lung anatomy, other internal or external non-anatomical objects are likely to be present in CXR images taken from ICU patients, including items such as endotracheal and nasogastric tubes, peripherally inserted central catheter lines, as well as other medical devices such as electrodes and cardiac pacemakers \citep{hunter2004medical}.

Table \ref{tab:results1} also shows the benefit of the lateral model, which achieves a larger average AUC ({\it 0.706}) compared to PA and AP frontal models. Per class, the Lateral model results in larger AUC values for the following findings: {\it consolidation}, {\it edema}, {\it effusion}, {\it hernia}, {\it mass}, {\it pneumonia} and {\it pneumothorax}.

\begin{table}[htbb]
  \centering 
  \caption{Per class AUC results of frontal and lateral baseline models on 13 thorax diseases and no finding.}
  \begin{tabular}{lccc}
    \hline
    {\bf Finding} & {\bf PA} & {\bf AP} & {\bf Lateral}\\
    \hline
    Atelectasis			& 0.760		& 0.666		& 0.753\\
    Cardiomegaly		& 0.868		& 0.746		& 0.794\\
    Consolidation		& 0.637		& 0.601		& 0.644\\
    Edema				& 0.736		& 0.745		& 0.766\\
    Effusion			& 0.719		& 0.740		& 0.754\\
    Fibrosis				& 0.706		& 0.643		& 0.666\\
    Hernia				& 0.731		& 0.742		& 0.788\\
    Infiltration			& 0.779		& 0.525		& 0.731\\
    Mass				& 0.646		& 0.624		& 0.680\\
    No Finding			& 0.742		& 0.677		& 0.765\\
    Nodule				& 0.534		& 0.603		& 0.556\\
    Pleural Thickening		& 0.688		& 0.620		& 0.660\\
    Pneumonia		 	& 0.611 		& 0.602		& 0.635\\
    Pneumothorax		& 0.672 		& 0.631		& 0.686\\
    \hline
    Average			& 0.702		& 0.655		& 0.706\\
    \hline    
  \end{tabular}
  \label{tab:results1} 
\end{table}

\subsection{DualNet Architecture Evaluation}

To evaluate the DualNet architecture, Table \ref{tab:results2} compares the per class AUC results for the subset of radiological studies in the MIMIC-CXR dataset where both frontal and lateral images were obtained. 
Both PA \& Lateral, as well as AP \& Lateral combinations are considered. First, we evaluate the performance of applying the separately trained frontal and lateral models to the images from these studies. We compare the individual model performance to the jointly trained DualNet architecture. In 12 out of 14 cases, the DualNet architecture performs better than applying separately trained models for PA \& Lateral studies\footnote{Note that for {\it pleural thickening} an improvement was witnessed beyond the scale shown in Table \ref{tab:results2}.}. For radiology studies that obtained AP \& Lateral radiographs, the DualNet architecture outperforms individual model classification for 10 of the 14 classes (highlighted in bold).

Overall, it can be seen that average AUC is greater for DualNet classifiers, compared to individually trained classifiers. For PA \& Lateral studies, DualNet achieves an average AUC of {\it 0.721} compared to {\it 0.690} for individually trained classifiers. For AP \& Lateral studies, DualNet achieves an average AUC of {\it 0.668} compared to {\it 0.637} for individually trained classifiers.

\begin{table}[htbb]
  \centering 
  \caption{Per class AUC results comparing DualNet to individual baseline classifiers for a subset of radiological studies where both frontal and lateral images were acquired.}
  \begin{tabular}{|l|cc|cc|}
    \hline
    {\bf Finding} & {\bf\makecell{Individual\\PA+Lateral}} & {\bf\makecell{DualNet\\PA+Lateral}} & {\bf\makecell{Individual\\AP+Lateral}} & {\bf\makecell{DualNet\\AP+Lateral}}\\
    \hline
    Atelectasis			& 0.760		& {\bf 0.766}		& {\bf 0.675}		& 0.671\\
    Cardiomegaly		& {\bf 0.835}	& 0.840			& 0.752			& {\bf 0.755}\\
    Consolidation		& {\bf 0.642}	& 0.632			& {\bf 0.625}		& 0.623\\
    Edema				& 0.723		& {\bf 0.734}		& {\bf 0.757}		& 0.749\\
    Effusion			& 0.735		& {\bf 0.757}		& 0.701			& {\bf 0.733}\\
    Fibrosis				& 0.638		& {\bf 0.761}		& 0.552			& {\bf 0.610}\\
    Hernia				& 0.716		& {\bf 0.815}		& 0.701			& {\bf 0.758}\\
    Infiltration			& 0.746		& {\bf 0.748}		& 0.590			& {\bf 0.773}\\
    Mass				& 0.656		& {\bf 0.692}		& 0.574			& {\bf 0.581}\\
    No Finding			& 0.746		& {\bf 0.758}		& 0.727			& {\bf 0.734}\\
    Nodule				& 0.527		& {\bf 0.568}		& {\bf 0.549}		& 0.527\\
    Pleural Thickening		& 0.687		& {\bf 0.687}		& 0.571			& {\bf 0.629}\\
    Pneumonia		 	& 0.596 		& {\bf 0.625}		& 0.571			& {\bf 0.593}\\
    Pneumothorax		& 0.659 		& {\bf 0.706}		& 0.577			& {\bf 0.621}\\
    \hline
    Average			& 0.690		& {\bf 0.721}		& 0.637			& {\bf 0.668}\\
    \hline
  \end{tabular}
  \label{tab:results2} 
\end{table}

\section{Conclusions and Future Work}

We have presented a collection of deep convolutional neural networks trained on the largest released dataset of chest x-ray images -- the MIMIC-CXR dataset. We evaluated our models on the task of recognizing the presence or absence of common thorax diseases. Separate models were trained to assess frontal, as well as lateral CXR inputs and a novel DualNet architecture was introduced that emulates routine clinical practice by taking into account both view types simultaneously. In future work, we plan to overcome several limitations of the current approach. First, several improvements could be made to our CNN training procedure, including the addition of techniques known to improve image-based classification performance, such as data augmentation and pixel normalization. More importantly, our models as currently described, only consider radiograph pixel information when making a classification decision. For several conditions (e.g. pneumonia), careful consideration of a patient's history and current clinical record is required to make an accurate final assessment. In future work we plan to incorporate this information within our model architecture. Finally, while automated radiograph analysis has many potential benefits, further consideration must be given to how these systems can best fit into clinical practice to aid workflow and be helpful for clinicians and the care of their patients.

\acks{The authors gratefully acknowledge Alistair Johnson from the MIT Laboratory for Computational Physiology for making the MIMIC-CXR dataset available, as well as Yifan Peng from the Biomedical Text Mining Group, Computational Biology Branch, NIH, for supplying the NegBio library.}


\appendix

\end{document}